\documentclass[letterpaper, 10 pt, conference]{ieeeconf}  

\IEEEoverridecommandlockouts                              

\makeatletter
\let\NAT@parse\undefined
\makeatother
\usepackage[pagebackref=false,breaklinks=true,colorlinks=true,bookmarks=false,linkcolor={red!50!black},urlcolor={blue},citecolor={green!60!black}]{hyperref} 
\usepackage{url}
\usepackage{pifont}
\usepackage{booktabs} 

\usepackage{etoolbox}
\makeatletter
\patchcmd{\@makecaption}
  {\scshape}
  {}
  {}
  {}
\makeatletter
\patchcmd{\@makecaption}
  {\\}
  {.\ }
  {}
  {}
\makeatother

\usepackage{graphicx}
\usepackage{subfigure}
\usepackage{lipsum}
\usepackage{cite}
\usepackage{bm}
\usepackage{esvect}
\usepackage{amsmath}
\usepackage[ruled,vlined]{algorithm2e}
\usepackage{color}
\usepackage{amsmath,amssymb}
\usepackage{tabularx}
\usepackage{multirow}
\usepackage{arydshln}
\usepackage[dvipsnames]{xcolor}

\usepackage{overpic}
\usepackage{overpic} 
\usepackage{color}
\usepackage{mathtools} 
\usepackage{currfile}
\usepackage{multirow}

\definecolor{turquoise}{cmyk}{0.65,0,0.1,0.3}
\definecolor{purple}{rgb}{0.65,0,0.65}
\definecolor{dark_green}{rgb}{0, 0.5, 0}
\definecolor{orange}{rgb}{0.8, 0.6, 0.2}
\definecolor{red}{rgb}{0.8, 0.2, 0.2}
\definecolor{darkred}{rgb}{0.6, 0.1, 0.05}
\definecolor{blueish}{rgb}{0.0, 0.3, .6}
\definecolor{light_gray}{rgb}{0.7, 0.7, .7}
\definecolor{pink}{rgb}{1, 0, 1}
\definecolor{greyblue}{rgb}{0.25, 0.25, 1}

\renewcommand{\paragraph}[1]{\vspace{.5em}\noindent\textbf{#1}.}

\usepackage{lipsum}

\usepackage{blindtext}

\newcommand{\kostas}[1]{{\color{Bittersweet} {[\bf Kosta: #1]}}}
\newcommand{\andrew}[1]{{\color{BlueViolet} {[Andrew: #1]}}}
\newcommand{\vitto}[1]{{\color{OrangeRed} {[Vitto: #1]}}}
\newcommand{\JB}[1]{{\color{OliveGreen} {[Jon: #1]}}}
\newcommand{\tom}[1]{{\color{RoyalPurple} {[Tom: #1]}}}
\newcommand{\pratul}[1]{{\color{Emerald} {[Pratul: #1]}}}
\newcommand{\at}[1]{{\color{blueish}#1}}
\newcommand{\AT}[1]{{\color{blueish}{\bf [Andrea: #1]}}}
\newcommand{\At}[1]{\marginpar{\tiny{\textcolor{blueish}{#1}}}}
\newcommand{\al}[1]{\textbf{\color{orange}[AL: #1]}}
\renewcommand{\kostas}[1]{}
\renewcommand{\andrew}[1]{}
\renewcommand{\vitto}[1]{}
\renewcommand{\JB}[1]{}
\renewcommand{\tom}[1]{}
\renewcommand{\pratul}[1]{}
\renewcommand{\at}[1]{}
\renewcommand{\AT}[1]{}
\renewcommand{\At}[1]{}
\renewcommand{\al}[1]{}

\makeatletter
\usepackage{xspace}
\DeclareRobustCommand\onedot{\futurelet\@let@token\@onedot}
\def\@onedot{\ifx\@let@token.\else.\null\fi\xspace}

\makeatother

\title{\LARGE \bf
HERO-SLAM: Hybrid Enhanced Robust Optimization of Neural SLAM
}

\author{Zhe Xin$^1$, Yufeng Yue$^2$, Liangjun Zhang$^3$, and Chenming Wu$^{3,\dagger}$ 
\thanks{$^{1}$Z. Xin is with the Institute of Automation, Chinese Academy of Sciences, Beijing, China.
        {\tt\small zhexin2015@gmail.com}}%
\thanks{
$^{2}$Y. Yue is with the School of Automation, Beijing Institute of Technology, Beijing, China.
{\tt\small yueyufeng@bit.edu.cn}}%
\thanks{
$^{3}$L. Zhang and C. Wu are with the Robotics and Autonomous Driving Lab (RAL), Baidu Research.
{\tt\small liangjun.zhang@gmail.com, wcm94@live.com}, $\dagger$ denotes corresponding author.}
\thanks{This work was partially supported by National Natural Science Foundation of China under Grant No. NSFC 62233002, 92370203. The authors would like to thank F. Zhang for the suggestions on figures.}
}

\begin{document}

\maketitle
\thispagestyle{empty}
\pagestyle{empty}

\begin{abstract}
Simultaneous Localization and Mapping (SLAM) is a fundamental task in robotics, driving numerous applications such as autonomous driving and virtual reality. Recent progress on neural implicit SLAM has shown encouraging and impressive results. However, the robustness of neural SLAM, particularly in challenging or data-limited situations, remains an unresolved issue. This paper presents HERO-SLAM, a Hybrid Enhanced Robust Optimization method for neural SLAM, which combines the benefits of neural implicit field and feature-metric optimization. This hybrid method optimizes a multi-resolution implicit field and enhances robustness in challenging environments with sudden viewpoint changes or sparse data collection.  Our comprehensive experimental results on benchmarking datasets validate the effectiveness of our hybrid approach, demonstrating its superior performance over existing implicit field-based methods in challenging scenarios. HERO-SLAM provides a new pathway to enhance the stability, performance, and applicability of neural SLAM in real-world scenarios.
Code is available on the project page: \href{https://hero-slam.github.io}{\textit{https://hero-slam.github.io}}.
\end{abstract}

\section{Introduction}
Visual Simultaneous Localization and Mapping
(SLAM) is a fundamental task in robotics and computer vision that drives many applications, spanning from the intricacies of robot navigation and 3D scene reconstruction, to the cutting-edge fields of autonomous driving and virtual reality. The essence of visual SLAM lies in its ability to reconstruct the structure and visual details of a 3D environment, all while tracking the camera's position in real-time. The keys to its success in real-world applications are relying on runtime efficiency, scalability, and most importantly, robustness.

\begin{figure}[t]
    \centering
    \includegraphics[width=0.90\linewidth]{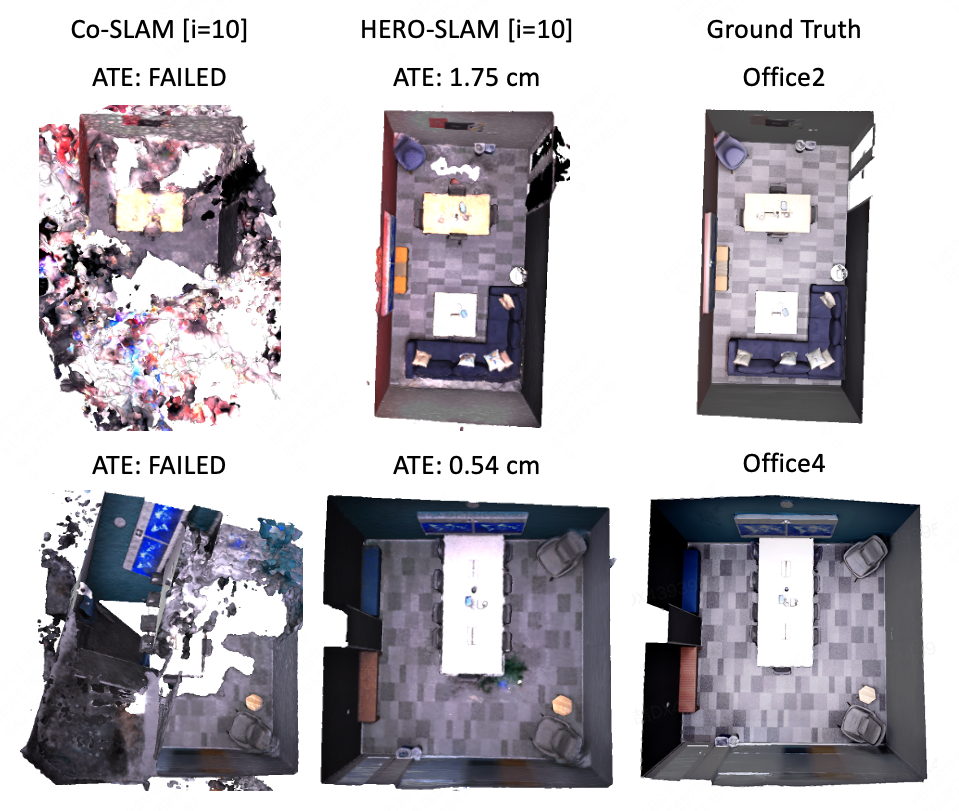}\vspace{-2mm}
    \caption{The visualization of mapping and tracking errors on Replica~\cite{replica-dataset} of challenging sparse inputs with large motion changes. This paper introduces a robust system for real-time dense 3D reconstruction, dubbed HERO-SLAM, which synergistically leverages the capabilities of neural implicit fields and feature-metric optimization, demonstrating exceptional resilience to large viewpoint changes and ensuring efficient runtime performance.}
    \label{fig:teaser}
\end{figure}

Visual SLAM can primarily be divided into two categories, sparse and dense, based on the nature of the map reconstruction. Specifically, sparse SLAM predominantly concentrates on deducing the camera trajectory from the sequential sensor data, generating sparse point clouds. In contrast, dense SLAM not only contemplates pose estimation but also initiates a detailed surface reconstruction.
Conventional dense visual SLAM approaches heavily lean on manually engineered features and matching strategies. These methods often incur a significant computational expense to solve pre-established optimization issues.
Recent advances in coordinate-based neural networks motivate many studies using implicit field representation in visual SLAM. A coordinate point can be encoded using sinusoidal positional or other formats of frequency encoding~\cite{vaswani2017attention} to represent high-frequency details compactly.
The benefits of using implicit field-based representation in dense visual SLAM tasks have been confirmed by pioneering work such as iMAP~\cite{sucar2021imap} and NICE-SLAM~\cite{zhu2022nice}. However, these methods are associated with high computational burdens and their running speeds are approximately $0.1$ to $1$ Hz, which restricts their applicability to a broader range of tasks.
Recent methods like Co-SLAM~\cite{wang2023co} and E-SLAM~\cite{johari2023eslam} are designed to push forward the boundary of implicit field-based visual SLAM. Compared to iMAP~\cite{sucar2021imap} and NICE-SLAM~\cite{zhu2022nice}, these methods have substantially improved the quality of dense reconstruction and pose estimation.
Despite these improvements, an important issue that hinders the wider range application of neural SLAM is the robustness to tackle challenging scenes, for example, the circumstance when the number of provided frames falls below the standard camera frequency, which is very common in real-world applications due to the constraints such as limited bandwidth for data transferring or storage availability. Under these conditions, the success rate of existing methods is not satisfactory.
In short, while recent advancements in neural implicit field-based visual SLAM have shown promise, there is still a need to improve their robustness and applicability in real-world applications.

The robustness issue that exists in neural SLAM approaches comes from the difficulty of optimizing neural networks. Despite the diverse underlying neural representations used to describe the implicit fields - including multi-layer perceptron (MLP)~\cite{mildenhall2021nerf}, hash grid~\cite{mueller2022instant}, codebook~\cite{takikawa2022variable}, triplane~\cite{chan2022efficient}, dense grid~\cite{sun2022direct}, 3D Gaussian~\cite{li2024GGRt} - they all essentially function as large nonlinear optimization systems. Therefore, input images' quality, view coverage, and relevance are key determinants of the neural implicit fields. However, in situations where data is challenging or limited, the low relevance across all data frames can easily mislead the optimization process toward ambiguous local solutions. 
In light of these challenges, our work seeks to explore a new path, which in particular designs a \textbf{\emph{hybrid}} representation that leverages both the capabilities of neural implicit field and feature-metric optimization. We aim to address the robustness problem for dense neural SLAM. This approach significantly improves the stability and performance of SLAM methods, particularly in challenging and data-limited situations, as shown in Fig.~\ref{fig:teaser}.
The contributions of our work are summarized as follows.


\begin{itemize}
    \item We propose a method that effectively leverages the advantages of neural implicit field and feature-metric optimization for visual SLAM. This results in increased robustness, especially in challenging environments involving abrupt view changes or sparse data collection.
    \item We propose a novel pipeline to optimize the hybrid feature-metric implicit fields using multiscale patch-based loss, which computes based on the warpings between feature points, feature maps, and RGB-D pixels.
    \item The comprehensive experiments on widely used benchmark datasets validate the effectiveness of our hybrid approach, particularly its superior performance compared to existing neural implicit field-based methods in challenging scenarios.
\end{itemize}
The outline of this paper is as follows. Section \ref{sec:related_work} provides a comprehensive literature review. We then present the detailed illustration of our proposed method, HERO-SLAM, in Section \ref{sec:hero_slam}. In Section \ref{sec:experiment}, we extensively evaluate the performance of our method and validate the effectiveness of its various modules.








\section{Related Work}
\label{sec:related_work}
\subsection{Visual SLAM}

Visual SLAM has emerged as a fundamental research area in the domains of robotics and computer vision. 
Traditional approaches for sparse/semi-dense visual SLAM include MonoSLAM~\cite{davison2007monoslam}, ORB-SLAM~\cite{mur2015orb}, VINS~\cite{qin2018vins}, LSD~\cite{engel2014lsd}, DSO~\cite{wang2017stereo}, where feature matching is widely used to recover the camera poses. In dense visual SLAM area, traditional approaches include KinectFusion~\cite{newcombe2011kinectfusion}, ElasticFusion~\cite{whelan2015elasticfusion}, where RGB-D sensors are required and the scene completeness is unsatisfying. Recently, deep learning has gained more and more attention, Droid-SLAM~\cite{teed2021droid} estimates motion fields between frames, which is highly computationally expensive and requires a large memory footprint. TANDEM~\cite{koestler2022tandem} uses a pre-trained MVSNet\cite{yao2018mvsnet}-like neural network on monocular depth estimation. Unlike these methods, our work uses neural implicit fields to estimate camera poses and reconstruct the scene simultaneously, achieving better scene completeness and higher rendering quality for less observed regions.





\subsection{Neural Implicit Field SLAM}
Neural implicit fields have become a significant area of research in computer vision and robotics, offering a novel paradigm for scene representation and reconstruction.
The Implicit Mapping and Planning (iMAP) framework~\cite{sucar2021imap} pioneers the use of deep implicit functions to represent 3D environments, providing a foundation for subsequent studies. The Neural Radiance Fields (NeRF) based LOAM (NeRF-LOAM)~\cite{deng2023nerf} extends this work by incorporating LiDAR odometry, enabling more accurate and efficient mapping. NICE-SLAM~\cite{zhu2022nice} and Co-SLAM~\cite{wang2023co} further expand on this by proposing improvements in efficiency and scalability, respectively. DIM-SLAM~\cite{li2023dense} and E-SLAM~\cite{johari2023eslam} demonstrate the versatility of neural implicit fields, showing how they can be used for dynamic scene reconstruction and event-based vision, respectively. 
Our work stands on the shoulders of these successful approaches, enhancing the robustness of existing neural implicit field SLAM methods.







\begin{figure*}[t]
    \centering
    \includegraphics[width=\linewidth]{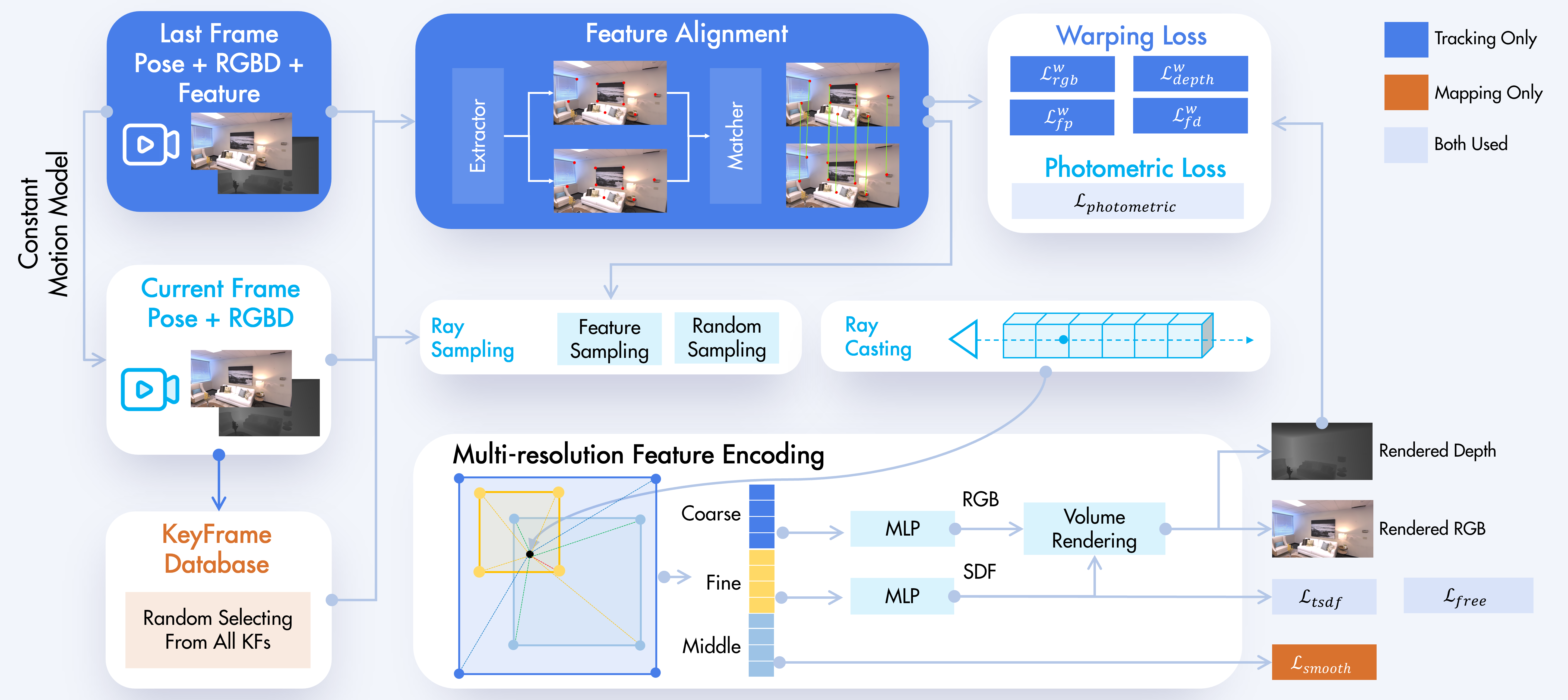}\vspace{-2mm}
    \caption{The overview of HERO-SLAM. We use hybrid optimization to enhance the robustness of neural SLAM. Every newly captured frame would be aligned with the last frame for the camera pose estimation using feature-metric warping losses. The robustness and accuracy of tracking get improved, which in turn, facilitates the enhancement of mapping quality by optimizing the neural implicit field of multi-resolution feature encoding. The mapping module optimizes all keyframes from the keyframe database based on photometric reconstruction and depth supervision, following the volumetric rendering paradigm.}
    \label{fig:pipeline}
\end{figure*}

\subsection{Pose Optimization within NeRF}
Another strand of research that bears similarity to our work pertains to the pose optimization within NeRF. However, the problem configuration diverges slightly from ours, as these methodologies do not necessitate the use of temporal data.
The body of literature is growing, with key works including ~\cite{wang2021nerf}, which presents a novel approach to jointly optimize camera poses and scene representations. This has been further developed by the Bundle-Adjustable Radiance Field (BARF)~\cite{lin2021barf}, which integrates the bundle adjustment approach for more accurate and flexible 3D reconstructions.
\cite{jeong2021self} makes a significant contribution by proposing a self-calibration mechanism for pose optimization, enhancing the accuracy of generated views. 
NoPe-NeRF~\cite{bian2023nope} presents a novel approach for pose estimation using the neural implicit fields, which has important implications for pose optimization in NeRF. 
LocalRF~\cite{meuleman2023progressively} introduces a progressive optimization strategy to improve the robustness of view synthesis. It is worth noting that these methods primarily aim at reconstructing large-scale scenes. Consequently, the optimization process usually takes place offline and is associated with significant time expenditure.

\section{HERO-SLAM}
\label{sec:hero_slam}

\subsection{Overview}
An overview pipeline of HERO-SLAM is shown in Fig.~\ref{fig:pipeline}. The architecture of our SLAM system is similar to traditional dense SLAM systems, which has a tracking module to recover the pose of each frame, and a mapping module to reconstruct dense scenes from the tracked frames. 
We utilize a multi-resolution grid as the representation of the spatial feature, which can approximate an implicit function that encodes the geometry and visual appearance of the scene.
Through the process of sampling features from the volumetric grid along the viewing rays and subsequently querying these sampled features with the Multi-Layer Perceptron (MLP) decoder, we can use learning-based optimizers to optimize the rendering of each pixel's color and depth based on inferred camera parameters in a differentiable manner.

Our proposed system takes in a sequence of RGB-D frames over time, with varying data intervals and motion between frames. This sometimes results in challenging scenes for existing neural implicit field SLAM approaches but commonly happens in real-world applications. Our work advances the robustness of neural SLAM by proposing a hybrid enhanced robust optimization scheme, enabling us to leverage the neural SLAM in a variety of environments, achieving high-quality pose recovery and dense mapping.

\subsection{Neural Implicit Field}
\label{sec:neural_implcit_field}
\subsubsection{Multi-resolution Neural Representation}

A multi-resolution grid for implicit functions provides flexible and scalable means to encode complex geometrical and topological information. The grid's varying resolution allows for a greater level of detail where required, effectively capturing intricate aspects of the implicit function, while conserving computational resources in less detailed regions. 
The grid is used to encode the implicit functions representing the geometry of the 3D scene. The implicit function $f_{\theta}(\mathbf{x})$ at a location $\mathbf{x}$ in 3D space is represented as an MLP with parameters $\theta$, which is trained to predict SDF values and appearance colors.

\subsubsection{Color, Depth and Truncated Signed Distance} Folloing~\cite{sucar2021imap, zhu2022nice}, the representation of a scene can be effectively undertaken by employing multi-resolution representation with MLPs. A function in three dimensions, which accepts a spatial location $\mathbf{x}=(x, y, z)$ as an input, can be utilized to represent the scene:

\begin{equation}
\sigma, {\rm \bf c} = f_\theta({\mathbf{x}}).
\end{equation}
when given a set of images ${{I}_i}$ and the estimated poses ${P_i}$, we can sample particles to describe the intensity of the light that is either blocked or emitted along the ray. The color $\widehat{\mathcal{C}}({\rm{\bf r}})$ and depth $\widehat{\mathcal{D}}({\rm{\bf r}})$ of a ray $\rm \bf r$ can be approximated by integrating the sampled particles along the ray as follows:
\vspace{-1mm}
\begin{equation}
\widehat{\mathcal{C}}({\rm{\bf r}}) = \sum_{i=1}^N {\rm exp}(-\sum_{j=1}^{i-1}\sigma_i\delta_i) (1-{\rm exp}(-\sigma_i\delta_i)) \rm {\bf c}_i ,
\label{nerf:color}
\end{equation}

\begin{equation}
\widehat{\mathcal{D}}({\rm{\bf r}}) = \sum_{i=1}^N {\rm exp}(-\sum_{j=1}^{i-1}\sigma_i\delta_i) (1-{\rm exp}(-\sigma_i\delta_i)) \sum_{j=1}^i\delta_j,
\label{nerf:depth}
\end{equation}
where ${\rm exp}(-\sum_{j=1}^{i-1}\sigma_i\delta_i)$ represents the accumulated transmittance along the ray from the first sample to the $i$-th sample. The term $(1-{\rm exp}(-\sigma_i \delta_i))$ denotes the alpha value of the current sample contributing to the rendered color and depth, while $\sigma_i$ is the density of sample $i$, $c_i$ is the predicted color of sample $i$, and $\delta_i$ is the distance from sample $i$ to its next sample $i+1$. 
To supervise the training of $f_\theta$, an $L_2$ photometric reconstruction loss is used:

\begin{equation}
    \mathcal{L}_{photometric}= \sum_i {\mathop{\mathbb{E}}_{{\rm \bf r}\in I_i} {|| \widehat{\mathcal{C}}({\rm{\bf r}}) - \mathcal{C}^{gt}_i({\rm{\bf r}}) ||_2^2}},
\label{eq:photometric}
\end{equation}
where $\mathcal{C}^{gt}_i({\rm{\bf r}})$ is the ground truth color of $\rm \bf r$ from image $I_i$. We follow~\cite{wang2023co} to make a conversion from the density to the Truncated Signed Distance Field (TSDF) $s_i$ by $\sigma_i=1/((1+\exp^{s_i/\epsilon})\cdot (1+\exp^{-s_i/\epsilon}))$, where $\epsilon$ is the parameter to truncate the distance field.
Likewise, the depth supervision is applied to the TSDF as follows.
\vspace{-3mm}

\begin{equation}
\mathcal{L}_{tsdf}= \sum_i {\mathop{\mathbb{E}}_{{\rm \bf r}\in I_i, |\mathcal{D}^{gt}_i({\rm{\bf r})| < \epsilon}} {|| \widehat{\mathcal{D}}({\rm{\bf r}}) - \mathcal{D}^{gt}_i({\rm{\bf r}}) ||_2^2}}  
.
\label{eq:depth}
\end{equation}
Besides, we adopt the same free space and smooth supervision from~\cite{wang2023co} to formulate $\mathcal{L}_{free}$ and $\mathcal{L}_{smooth}$.


\subsection{Hybrid Enhanced Robust Optimization}

The optimization scheme in Sec.~\ref{sec:neural_implcit_field} is computed in pixel-wise, while the underlying relationship in spatial-wise is not explicitly supervised by any loss function. We argue that this optimization scheme easily fails when relative motion between two consecutive frames is large, as the fact that the learning-based optimization uses the gradient descent method, which heavily relies on the initial guess and is easily stuck at local minima.
Drawing the inspiration from~\cite{zheng2014patchmatch, davison2007monoslam,wu2023mapnerf}, we additionally impose explicit supervision among the pairs of frames to facilitate tracking and mapping, by homography warping.
Our hybrid enhanced robust optimization extends the neural implicit field SLAM from the perspective of feature metric matching, and all concluded into a set of warping losses to strengthen the supervision among different frames.

We first extend the pixel-wise photometric and depth supervision (Eq.~\ref{eq:photometric} and~\ref{eq:depth}) to multi-frame under the assumption of reprojection transformation obtained from the tracking module. To conquer the accuracy issue of warping, we adopt Structural Similarity Index (SSIM)~\cite{wang2004image} with $3\times 3$ patches to compute $\mathcal{L}^w_{rgb}$ and $\mathcal{L}^w_{depth}$. Considering two frames $I_i$ and $I_j$, the relative transformation in between is denoted as $\mathbf{R}_{i}^j$ and $\mathbf{t}_{i}^j$, we denote $\mathbf{H}_{ji}$ be the homography between them and $\mathbf{K}$ is the intrinsic parameter of the used camera. 
\begin{equation}
    \mathbf{H}_{ji} = \mathbf{K}( \mathbf{R}_i^j + \frac{\mathbf{t}_i^j \mathbf{n}_i^T \mathbf{R}_i^T}{\mathbf{n}_i^T(\mathbf{q}_i +\mathbf{R}^T_i \mathbf{t}_i)}),
\end{equation}
where $\mathbf{n}_i$ is the normal. We denote a patch $\mathbf{P}_{\mathbf{q}_i}$ as the $3\times 3$ patch centered at $\mathbf{q}_i$ in $I_i$, then its warped patch $\mathbf{P}_{\mathbf{q}_j}$ can be obtained by $\mathbf{H}_{ji} \mathbf{P}_{\mathbf{q}_i}$. We introduce the formulation of visibility mask $M_i(\mathbf{P}_{\mathbf{q}_i})$ from~\cite{darmon2022improving} to avoid warp invisible patches. Then we can define the warping losses of color and depth as follows.

\begin{equation}
    \mathcal{L}^w_{rgb} = \frac{\sum_i M_i \cdot \text{SSIM}(I_i(\mathbf{P}_i), I_j(\mathbf{H}_{ji} \mathbf{P}_i))}{\sum_i M_i}  
\label{eq:photometric_warp}
\end{equation}

\begin{equation}
        \mathcal{L}^w_{depth} = \frac{\sum_i M_i \cdot \text{SSIM}(D_i(\mathbf{P}_i), D_j(\mathbf{H}_{ji} \mathbf{P}_i))}{\sum_i M_i}  
\label{eq:depth_warp}
\end{equation}

To further optimize the robustness of the proposed system, 
we opt to use the feature metric descriptor in image space to provide additional supervision to indicate the neural network optimized toward the guided direction.
We use SuperPoint~\cite{detone2018superpoint},
which is a deep learning-based method designed for joint detection and description of interest points in an image,
to extract the feature metric descriptor $F$ from each frame. 
Then we apply LightGlue~\cite{lindenberger2023lightglue} to match the feature maps $F_i$ and $F_j$, which augments the visual descriptors with context based on self- and cross-attention units with positional encoding. This helps introspect the feature maps and predicts a set of correspondences $\mathbf{S}_{ij}$ between the two frames, based on their pairwise similarity and unary matchability. Then, the projected pixel $\Pi(\mathbf{q}_i)$ in frame $I_j$ of a pixel $\mathbf{q}_i$ lifted from frame $I_i$ with homogeneous representation can be obtained by 
$\Pi(\mathbf{q}_i)=\mathbf{K} \left({D}_{\mathbf{q}_i}{\mathbf{R}}_{i}^j \mathbf{K}^{-1} \mathbf{q}_i+{\mathbf{t}}_{i}^j\right)
$, where ${D}_{\mathbf{q}_i}$ is the depth of $\mathbf{q}_i$.





To utilize the feature maps and the correspondences, we propose a hybrid enhanced robust optimization scheme, which optimizes the following feature points and feature maps pixel-wise loss functions during the tracking process,
\begin{equation}
    \mathcal{L}^w_{fp} = \frac{ \sum_{(\mathbf{q}_i, \mathbf{q}_j) \in S_{ij}} M_j ||\mathbf{q}_j - \Pi(\mathbf{q}_i))||_2}{\sum_{ij} M_j}
\end{equation}

\begin{equation}
    \mathcal{L}^w_{fd} = \frac{\sum_{(\mathbf{q}_i, \mathbf{q}_j) \in S_{ij}} M_j \cdot |F_j(\mathbf{q}_j) - F_j(\mathcal{q}(\mathbf{q}_i))|}{\sum_{ij} M_j}
\end{equation}

The overall loss function $\mathcal{L}$ in optimizing the neural SLAM is defined as the sum of the above loss functions.





\section{Experiments}
\label{sec:experiment}





\begin{table*}[ht]
\setlength{\tabcolsep}{2.4mm}
\renewcommand\arraystretch{1.1}
\caption{Quantitative results of all eight scenes on the Replica dataset~\cite{replica-dataset}. Our method achieves better reconstruction quality and has the best average performance in all metrics, even with low-frequency image sequences.} \vspace{-2mm} 
\label{tab:replica-res}
\hspace*{0cm}\makebox[\linewidth][c]{%
\begin{tabular}{ c  c  c c c c c  c c c c }
\toprule
\multirow{2}*{Metrics} & \multirow{2}*{Method} & \multicolumn{8}{c}{Replica} & \multirow{2}*{Avg.} \\
& & Office0 & Office1 & Office2 & Office3 & Office4 & Room0 & Room1 & Room2 & \\
\hline
\multirow{7}*{Depth L1[\textit{cm}]$\downarrow$}
& iMAP~\cite{sucar2021imap} & 3.79 & 3.76 & 3.97 & 5.61 & 5.71 & 5.08 & 3.44 & 5.78 & 4.64\\
& NICE-SLAM~\cite{zhu2022nice} & 1.43 & 1.58 & 2.70 & 2.10 & 2.06 & 1.79 & 1.33 & 2.20 & 1.90\\
& Vox-Fusion~\cite{yang2022vox} & 3.44 & 1.77 & 3.52 & 1.82 & 4.84 & 1.76 & 2.52 & 3.58 & 2.91\\
& Co-SLAM~\cite{wang2023co} & 1.24 & 1.48 & 1.86 & 1.66 & 1.54 & 1.05 & \textbf{0.85} & 2.37 & 1.51\\
& Co-SLAM[i=10] & 1.13 & 2.57 & 64.69 & 1.66 & 41.32 & 75.14 & 2.99 & 2.31 & \ding{56} \\
& Ours[i=10] & \textbf{1.12} & 1.47 & 1.85 & \textbf{1.52} & 1.54 & 0.93 & 0.99 & \textbf{2.18} & 1.46\\
& Ours[i=5] & 1.17 & \textbf{1.41} & \textbf{1.72} & 1.56 & \textbf{1.46} & \textbf{0.91} & \textbf{0.85} & \textbf{2.18} & \textbf{1.41}\\
\midrule
\multirow{7}*{Acc.[\textit{cm}]$\downarrow$}
& iMAP & 3.34 & 2.10 & 4.06 & 4.20 & 4.34 & 4.01 & 3.04 & 3.84 & 3.62\\
& NICE-SLAM & 1.85 & 1.56 & 3.28 & 3.01 & 2.54 & 2.44 & 2.10 & 2.17 & 2.37\\
& Vox-Fusion & 1.63 & 1.44 & 3.03 & \textbf{2.33} & \textbf{2.02} & \textbf{1.77} & \textbf{1.51} & 2.33 & 2.01 \\
& Co-SLAM & 1.57 & 1.31 & 2.84 & 3.06 & 2.23 & 2.11 & 1.68 & 1.99 & 2.10\\
& Co-SLAM[i=10] & 1.68 & 1.35 & 46.72 & 2.73 & 11.09 & 17.58 & 3.22 & 1.95 & \ding{56} \\
& Ours[i=10] & 1.52 & 1.28 & 2.65 & 2.80 & 2.28 & 2.01 & 1.63 & 1.90 & 2.01 \\
& Ours[i=5] & \textbf{1.51} & \textbf{1.26} & \textbf{2.55} & 2.58 & 2.23 & 1.97 & 1.53 & \textbf{1.87} & \textbf{1.94} \\
\midrule
\multirow{7}*{Comp.[\textit{cm}]$\downarrow$}
& iMAP & 3.62 & 3.62 & 4.73 & 5.49 & 6.65 & 5.84 & 4.40 & 5.07 & 4.93\\
& NICE-SLAM & 1.84 & 1.82 & 3.11 & 3.16 & 3.61 & 2.60 & 2.19 & 2.73 & 2.63\\
& Vox-Fusion & 1.87 & \textbf{1.44} & 3.03 & 2.81 & 3.51 & 2.69 & 2.31 & 2.58 & 2.53\\
& Co-SLAM & 1.56 & 1.59 & 2.43 & 2.72 & \textbf{2.52} & \textbf{2.02} & 1.81 & 1.96 & \textbf{2.08} \\
& Co-SLAM[i=10] & 1.61 & 1.77 & 11.19 & 2.74 & 15.47 & 17.10 & 3.13 & 2.08 & \ding{56} \\
& Ours[i=10] & 1.53 & 1.70 & 2.38 & \textbf{2.68} & 2.67 & 2.16 & 1.85 & 1.94 & 2.11\\
& Ours[i=5] & \textbf{1.50} & 1.62 & \textbf{2.34} & 2.70 & 2.60 & 2.20 & \textbf{1.78} & \textbf{1.93} & \textbf{2.08}\\
\midrule
\multirow{7}*{Comp. Ratio\%$\uparrow$}
& iMAP & 83.59 & 88.45 & 79.73 & 73.90 & 74.77 & 78.34 & 85.85 & 79.40 & 80.50\\
& NICE-SLAM & 94.93 & 94.11 & 88.27 & 87.68 & 87.23 & 91.81 & 93.56 & 91.48 & 91.13\\
& Vox-Fusion & 93.86 & 94.40 & 88.94 & 89.10 & 86.53 & 92.03 & 92.47 & 90.13 & 90.93\\
& Co-SLAM & 96.09 & \textbf{94.65} & 91.63 & 90.72 & \textbf{90.44} & \textbf{95.26} & 95.19 & 93.58 & 93.44 \\
& Co-SLAM[i=10] & 95.06 & 93.56 & 42.45 & 90.45 & 56.57 & 43.56 & 79.92 & 92.51 & 74.26\\
& Ours[i=10] & 96.20 & 94.02 & 92.08 & \textbf{91.13} & 89.56 & 94.57 & \textbf{95.54} & 93.41 & 93.31\\
& Ours[i=5] & \textbf{96.44} & 94.58 & \textbf{92.44} & 91.00 & \textbf{90.44} & 94.45 & 95.24 & \textbf{93.86} & \textbf{93.53}\\

\bottomrule
\end{tabular}}
\vspace{-2mm}
\end{table*}

\subsection{Experimental Setup}
\textbf{Datasets}. We use three datasets for evaluation. Synthetic Replica~\cite{replica-dataset} dataset is used to verify the quality of our reconstruction. Real-world ScanNet~\cite{scannet-dataset} and TUM RGB-D~\cite{tum-dataset} datasets are used to evaluate pose estimation. Each dataset provides ground truth pose data.
We follow~\cite{zhu2022nice} to pre-processing all testing data.

\begin{figure}[!bp]
    \includegraphics[width=\linewidth]{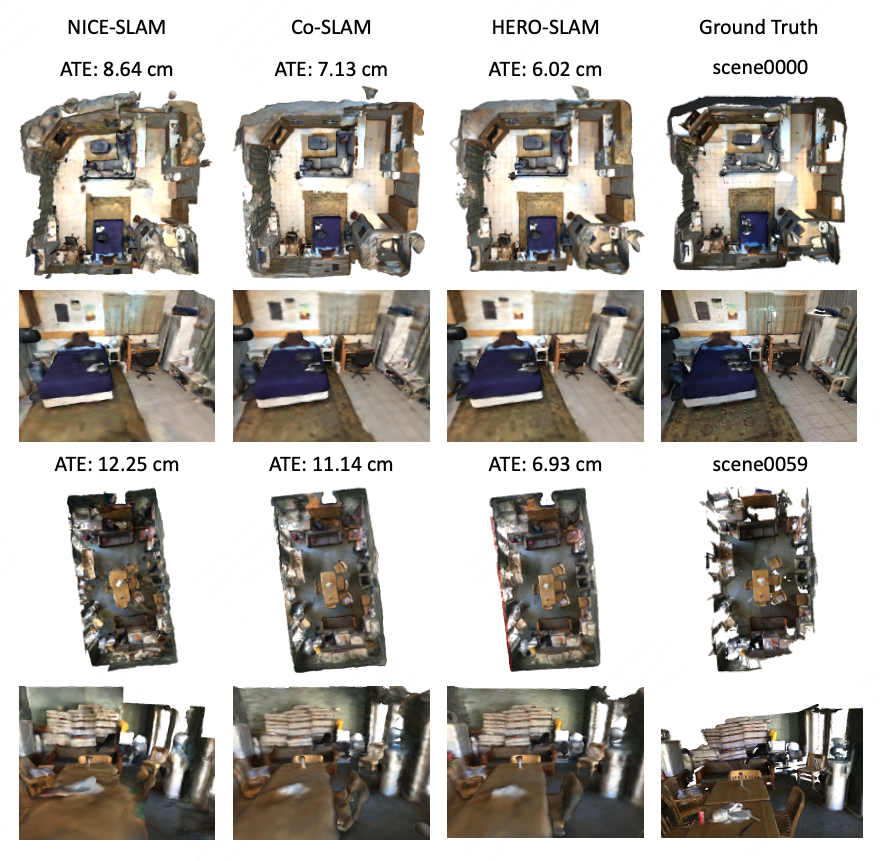}
    \caption{Qualitative visualization of results among different approaches. Our reconstructions are smoother, more complete, and have fewer artifacts compared to other advanced methods on the ScanNet dataset~\cite{scannet-dataset}.}
    \label{fig:compare}
\end{figure}

\textbf{Metrics}. To evaluate the quality of reconstruction, we utilize both 2D and 3D metrics. In the 2D metrics, we measure the Depth L1 (cm) by comparing the estimated and actual meshes at randomly selected points. On the other hand, in 3D, we evaluate the accuracy (cm), completion (cm), and completion ratio (\%) with a threshold of 5cm. To achieve this, we employ the mesh culling strategy in Co-SLAM~\cite{wang2023co}, which eliminates unobserved regions and noisy points outside the camera frustum and target scene. In terms of pose estimation evaluation, we use ATE RMSE (cm).

\textbf{Baselines}. We select several recent neural SLAM systems for comparison, iMAP~\cite{sucar2021imap}, NICE-SLAM~\cite{zhu2022nice}, DI-FUSION~\cite{di-fusion}, Go-SLAM~\cite{go-slam}, Vox-Fusion~\cite{yang2022vox} and Co-SLAM~\cite{wang2023co}. To examine the accuracy and quality at different image frequencies, results are presented using the notation `Method[i=n]', where n represents the interval between images. No suffix means using all images for testing.

\textbf{Implementation Details}. The experiments are conducted on a desktop PC with a 3.60GHz Intel Core i9-9900K CPU and an NVIDIA RTX 2080ti GPU. Tracking and mapping were performed using 100 iterations with an interval of 5 for all methods, and 200 iterations with an interval of 10. For tracking, 1,024 pixels are sampled, while 2,048 pixels are sampled for global optimization in all keyframes.

\subsection{Evaluation of Tracking and Mapping}
\noindent \textbf{Evaluation on Replica~\cite{replica-dataset}.}
Detailed comparison results of all eight scenes are shown in Tab.~\ref{tab:replica-res}. Despite using lower-frequency images, our proposed method outperforms the baselines in both 2D and 3D metrics. In contrast, methods like NICE-SLAM~\cite{zhu2022nice} and Co-SLAM~\cite{wang2023co} rely solely on the uniform motion model, which can easily cause tracking drift and ultimately lead to reconstruction failure. Our method improves the robustness of the neural SLAM system by building feature correspondences between the current and former frames. Texture and feature metric warping constraints are used to optimize the camera pose. Furthermore, even as the image frequency decreases (from $i=5$ to $10$), our method still achieves good results with only a slight decrease, and a 100\% success rate. On the contrary, Co-SLAM~\cite{wang2023co} fails to reconstruct several scenes at i=10, the average success rate is 62\%.
We present the evaluation of the reconstruction quality using the culling strategy from NICE-SLAM~\cite{zhu2022nice} in Tab.~\ref{tab:avg-replica}. Even with low image frequency, our method exhibits the best overall performance.

\begin{table}[t]
\caption{Reconstruction results on Replica dataset~\cite{replica-dataset} using NICE-SLAM~\cite{zhu2022nice} culling strategy (unit: cm).}  
\vspace{-2mm}
\label{tab:avg-replica}
\hspace*{0cm}\makebox[\linewidth][c]{%
\begin{tabular}{ c  c c c c }
\toprule
{Methods} & Depth L1$\downarrow$ & Acc.$\downarrow$ & Comp.$\downarrow$ & Comp. Ratio$\uparrow$ \\ 
\midrule
iMAP~\cite{sucar2021imap} & 7.64 & 6.95 & 5.33 & 66.60 \\ 
DI-FUSION~\cite{di-fusion} & 23.33 & 19.40 & 10.19 & 72.96 \\ 
NICE-SLAM~\cite{zhu2022nice} & 3.53 & 2.85 & 3.00 & 89.33 \\ 
Go-SLAM~\cite{go-slam} & 3.38 & 2.50 & 3.74 & 88.09 \\ 
Co-SLAM~\cite{wang2023co} & 1.58 & \textbf{2.15} & 2.21 & 92.99 \\ 
Ours[i=5] & \textbf{1.41} & 2.62 & \textbf{2.15} & \textbf{93.22} \\ 
Ours[i=10] & 1.46 & 2.73 & 2.14 & 93.13 \\ 

\bottomrule
\end{tabular}}
\end{table}

\begin{figure}[b]
\vspace{-5mm}
    \centering
    \includegraphics[width=\linewidth]{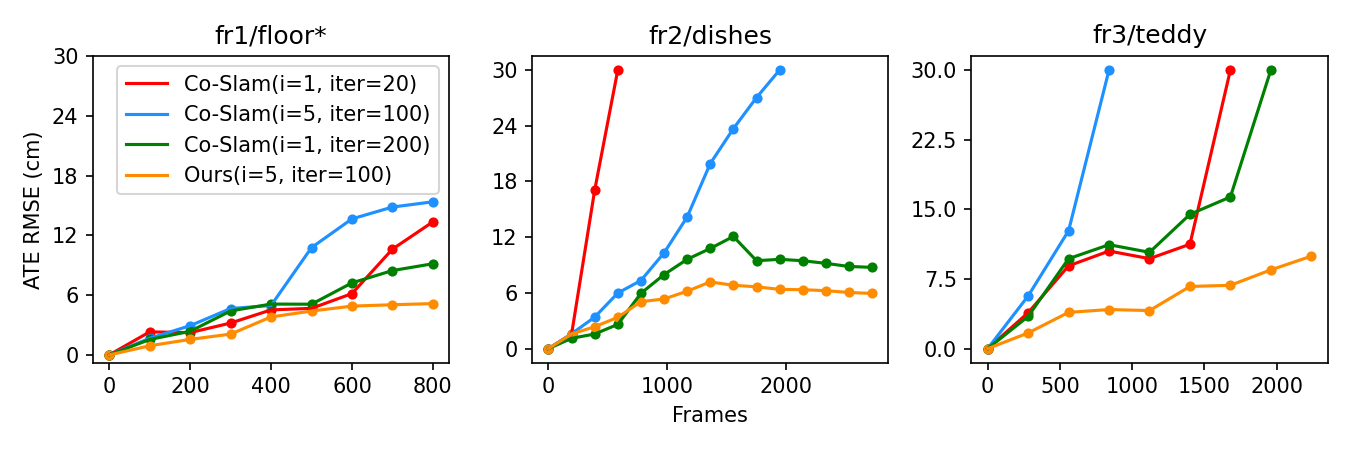}\vspace{-2mm}
    \caption{A comparison of ATE RMSE trends as the number of images increases across different scenes.}
    \label{fig:tum_res}
\vspace{-5mm}
\end{figure}

\textbf{Evaluation on TUM RGB-D~\cite{tum-dataset}.}
We compare the accuracy of pose estimation on the TUM dataset with NeRF-based RGB-D SLAM~\cite{wang2023co}. However, Co-SLAM~\cite{wang2023co} requires continuous images, the poses of some scenes in the TUM dataset may not be continuous. We only test using the first continuous segment of these scenes. According to Table ~\ref{tab:tum-res}, our method achieves the highest and most reliable tracking performance. The TUM dataset contains many hand-held shooting scenes with significant viewpoint changes during movement. Our algorithm can effectively handle such changes through feature-metric optimization. Although increasing tracking iterations can lead to better results, our method still outperforms Co-SLAM~\cite{wang2023co} by a large margin. Fig.~\ref{fig:tum_res} illustrates how our method mitigates cumulative error and pose drift facing large viewpoint changes in challenging scenarios as the number of images increases.

\begin{table}[t]
\centering
\renewcommand\arraystretch{1.1}
\caption{Camera tracking results on TUM RGB-D dataset~\cite{tum-dataset}. Our method achieves the best performance and is robust to large view changes. Non-continuous scenes are marked with an asterisk. Trajectories with errors larger than 30 cm are denoted as {FAILED} across the paper.}  \vspace{-2mm}
\label{tab:tum-res}
\hspace*{0cm}\makebox[\linewidth][c]{%
\begin{tabular}{ c c c c c c}
\toprule
ATE RMSE (\textit{cm}) & \multicolumn{3}{c}{Co-SLAM~\cite{wang2023co}} & Ours \\
\hline
Interval & $i = 1$ & $i = 5$ & $i = 1$ & $i = 5$ \\
\hline
Tracking Iters & iter = 20 & iter = 100 & iter = 200 & iter = 100 \\

\midrule
fr1/desk & \textbf{2.43} & {FAILED} & & 2.44 \\
fr1/floor$^{*}$ & 13.33 & 15.76 & 9.15 & \textbf{5.15} \\
fr2/desk$^{*}$ & {FAILED} & {FAILED} & 27.82 & \textbf{3.40} \\
fr2/dwp & {FAILED} & {FAILED} & 7.17 & \textbf{7.14} \\
fr2/dishes & {FAILED} & 24.02 & 8.75 & \textbf{6.24} \\
fr2/pslam$^{*}$ & {FAILED} & {FAILED} & {FAILED} & \textbf{11.12} \\
fr3/office & 2.40 & 2.40 & & \textbf{2.30}  \\
fr3/ntnwl & {FAILED} & 4.81 & 2.86 & \textbf{2.13} \\
fr3/teddy & {FAILED} & {FAILED} & {FAILED} & \textbf{9.95} \\
\bottomrule
\end{tabular}}
\vspace{-3mm}
\end{table}

\textbf{Evaluation on ScanNet~\cite{scannet-dataset}.}
We evaluate the tracking results on real-world sequences from ScanNet, where the ground-truth trajectories are generated using BundleFusion~\cite{dai2017bundlefusion}. In Fig.~\ref{fig:compare}, a qualitative analysis of $scene0000$ and $scene0059$ is presented. Our method achieves better pose accuracy compared to NICE-SLAM~\cite{zhu2022nice} and Co-SLAM~\cite{wang2023co}. Moreover, Fig.~\ref{fig:compare} shows that our method exhibits better reconstruction quality with smoother surfaces, consistent geometries, and fewer artifacts.

\subsection{Run-time and Performance Analysis}
Table~\ref{tab:time-memory} presents a comparison of run-time and performance for Replica~\cite{replica-dataset} Office2 and TUM-RGBD~\cite{tum-dataset} fr1/desk at different image frequencies. The run-time is measured in ms/iter $\times$ \#iter. Our method's tracking performance remains largely unaffected even with fewer tracking iterations. This showcases the resilience of our method to significant changes in perspective and its ability to operate in real-time.

\begin{table}[t]
\caption{Run-time, frame rate comparison, and pose estimation performance with different iterations when tracking.}  \vspace{-1mm}
\label{tab:time-memory}
\hspace*{0cm}\makebox[\linewidth][c]{%
\begin{tabular}{ c c c c c }
\toprule
\multirow{2}*{Method} & Track & Map & \multirow{2}*{FPS$\uparrow$} & ATE \\
 & (ms)$\downarrow$ & (ms)$\downarrow$ & & RMSE ( \textit{cm}) \\
 \midrule
 NICE-SLAM & 12.3$\times$50 & 125.3$\times$60 & 0.68 & - \\ 
 Co-SLAM & 7.8$\times$20 & 20.2$\times$10 & 6.4 & - \\ 
 \midrule
 Ours[i=10] & 11.3$\times$200 & 11.7$\times$200 & 4.43 & 1.75 \\ 
Ours[i=10] & 11.3$\times$100 & 11.7$\times$100 & 8.85 & 1.88 \\ 
\midrule
 Ours[i=5] & 11.3$\times$200 & 11.7$\times$200 & 2.22 & 2.37 \\ 
 Ours[i=5] & 11.3$\times$100 & 11.7$\times$100 & 4.43 & 2.44 \\

\bottomrule
\end{tabular}}
\end{table}

\subsection{Ablation Study}
We evaluate our feature-metric optimization by testing different warping loss combinations on tracking results, i.e., RGB and depth patch-wise warping, feature point, and feature map pixel-wise warping. Tab.~\ref{tab:ablation} reports the results tested on Replica~\cite{replica-dataset} and TUM-RGBD~\cite{tum-dataset}. In this experiment, we evaluate the absolute trajectory errors (ATE) without estimating the rigid transformation to align the estimated trajectory with the ground truth, as commonly done in traditional SLAM. As shown in Tab.~\ref{tab:ablation}, by using more warp losses, the trajectory tracking accuracy becomes better and better. Incorporating feature map pixel-wise warping losses into the TUM dataset has resulted in significant improvements in pose estimation. This is because detection and matching processes often encounter repeatability errors in feature points, which cannot be eliminated during pose optimization in tracking. Incorporating feature-metric supervision to maintain the semantic information's consistency around feature points is crucial for reducing errors. 

\begin{table}[t]
\setlength{\tabcolsep}{2.8mm}
\renewcommand\arraystretch{1.1}
\caption{Camera tracking performance evaluated using different combinations of warping losses, without trajectory alignment.}  \vspace{-1mm}
\label{tab:ablation}
\hspace*{0cm}\makebox[\linewidth][c]{%
\begin{tabular}{ c c c c | c c }
\toprule

\multicolumn{4}{c}{Warp Losses} & \multicolumn{2}{|c}{ATE RMSE (\textit{cm})} \\
RGB & Depth & KeyPoint & Feature & Office2 & fr1/desk \\ 
\midrule
\ding{56} & \ding{56} & \ding{56} & \ding{56} & {FAILED} & {FAILED} \\
\ding{52} & \ding{56} &  \ding{56} & \ding{56} & 4.28 & {FAILED} \\
\ding{52} & \ding{52} & \ding{56} & \ding{56} & 3.89 & 10.28 \\
\ding{52} & \ding{52} & \ding{52} & \ding{56} & 3.81 & 8.58 \\
\ding{52} & \ding{52} & \ding{52} & \ding{52} & 3.78 & 5.45 \\
\bottomrule
\end{tabular}}
\vspace{-5mm}
\end{table}

\section{Conclusion}
This paper presents HERO-SLAM, a hybrid optimization solution for neural SLAM that stands for Hybrid Enhanced Robust Optimization. By fusing the prowess of both neural implicit field and feature-metric optimization, our hybrid method optimizes
a multi-resolution implicit field and enhances robustness in
challenging environments with sudden viewpoint changes or
sparse data collection. The experimental results validate the effectiveness of our approach compared to existing methods, particularly in challenging scenarios.

\footnotesize

\clearpage

\bibliographystyle{IEEEtran} 
\bibliography{icra_abrv}

\end{document}